# Correlating Medi- Claim Service by Deep Learning Neural Networks.

**Jayanthi Vajiram [1], Negha Senthil [2] Nean Adhith.P [3]**

[1]Vellore Institute of Technology, Chennai, jayanthi.2020@vitstudent.ac.in(Teaching Research Associate)
[2]Vellore Institute of Technology, negha.2019@vitstudent.ac.in. (Law school)
[3]Valliammal Eng College, SRM, Chennai, namooshivnian23@gmail.com

**Abstract**
Organized crime is a continuous issue, and predicting it is always under research. Medical insurance claims are one of the organized crimes related to patients, physicians, diagnostic centers, and insurance providers, forming a chain reaction that must be monitored constantly. These kinds of frauds affect the financial growth of both the insured people and the health insurance companies. The Convolution Neural Network architecture is used to detect fraudulent claims through a correlation study of regression models, which helps to detect money laundering on different claims given by different providers. Supervised and unsupervised classifiers are used to detect fraud and non-fraud claims. By using different attributes of patient case studies, diagnostic reports, and service provider reimbursement claim codes as control variables and attributes of the target class to detect performance metrics, this paper highlights the top reason for organized crime through the public dataset. The claims are filed by the provider, so the fraud can be organized crime. The performance metrics of accuracy, sensitivity, specificity, recall, precision, AUC, and f1-scores are calculated. More features are extracted from the proposed model of Auto-Encoder, which gives 0.90 accuracy, 0.80 AUROC score, f1-score (0.61), and 0.55 Kappa Score. Different threshold values are used to detect a tradeoff between fraud and non-fraud class predictions.

# Introduction

Healthcare frauds are related to physicians, diagnostics centers, and insurance providers. These entities are used to deliver the costliest treatment procedures and drugs, often leading to bad practices including the claiming of more money than is due. As a result, insurance premiums and healthcare costs are high, particularly in low-income countries. Common types of fraud by providers include improper billing, duplicate billing of the same service, charging more than necessary for physician consultation fees and surgery, charging for expensive services, and billing for services that are not covered. Such frauds lead to patients receiving Mediclaim sanctions of only 70-80% of their hospitalization costs, leading to a financial crisis. This article explores the use of convolutional neural network architectures to analyze claims filed by providers, predicting whether they are fraudulent or not. Vyas's (2021) study showed that insurance fraud and their network analysis can be effectively detected using fraud detection systems in insurance claim systems [1]. Suma Latha's (2019) study on medical insurance fraud detection and medical claims utilized logistic regression for decision analysis and was found to be more effective than existing systems [2]. To classify providers into fraud and non-fraud categories, supervised and unsupervised models were used with feature selection, logistic regression, random forest, and auto-encoder. Classic logistic regression was used to predict the linearity of dependent and independent variables. Random forest model was used to handle large input variables with different dimensions of non-linearity. The proposed auto-encoder compressed the input variables into a latent space and then reconstructed the output variables. The model was trained on non-fraud data and then reconstructed it back as fraud data through reconstruction errors with a set threshold. The aim of this study is to accurately predict fraudulent providers by using correlation study of residual error from regression models, prediction based on linear and non-linear variables concerning threshold and intensity variance, supervised and unsupervised classifiers, feature extraction using principal component analysis and auto-encoders, and ensemble learning with hyperparameter tuning. The results of this study provide insights into the fraudulent patterns and claims, enabling a better understanding of future predictions of providers.

This study highlights (a) predicting fraudulent providers through a correlation study of residual error by regression models, (b) predicting based on linear and non-linear variables with regards to threshold and intensity variance, (c) using both supervised and unsupervised classifiers for model prediction, (d) utilizing principal component analysis for feature extraction, (e) calculating accurate predictions with a proposed Auto-Encoder model, and (f) increasing model performance with an ensemble learning method and hyperparameter tuning. This study provides insights into fraudulent patterns and claims, which can help to better predict future provider activities.

**Related study**

Wasson's (2019) study on organized crime among healthcare professionals aimed to reduce system costs and address future challenges [3]. Patel's study used period-based claim anomalies (k-means clustering) and disease-based (Gaussian distribution) anomalies outlier detection with a public dataset to analyze fraudulent insurance claims [4]. Daskalopoulou's (2019) study on healthcare providers' roles and levels of service delivery between patients and organizational level explained the new service provider roles [5]. Akbar's (2020) study analyzed fraud prediction using the statistical methods of random forest regression and extreme gradient boost (XGB) and compared recall and precision scores [6]. Saldamli's (2020) study on health insurance services covered accident or major illness and highlighted unsynchronized patient data and false claims. These were analyzed by the National health care anti-fraud association through blockchain fraud detection [7]. Powel's (8) study investigated healthcare providers' mental, work-related stress and protective factors. Mathew's (9) study explored the effects of low-income patients' claims and insurance companies delays and fewer refunds and the help of social services agencies support. Carcillo (2021) (10) examined the use of supervised and unsupervised learning techniques to improve credit card fraud detection accuracy. Madhavan's (11) study investigated the COVID-19 medical practices of direct and reverse polymerase chain reaction with Res-CovNet model analysis. Korneta's (12) study explored the effects of the COVID-19 pandemic time health care providers in Poland and their capitation payment scheme and delay settlement. Lastly, Mydin's (13) study analyzed healthcare service providers' attitudes and practice in managing elderly abuse and neglect cases and their negligence towards the elderly community medical claims. Haque et al. studied the identification of fraudulent records of patients' high Mediclaim insurance frauds and the insurance providers' loss due to frauds by using a mixture of clinical codes and a Long Short-Term Memory (LSTM+PCA) network with Robust Principal Component Analysis [15]. Sanober's study focused on fraud transactions and detection by Random Forest, SVM, Logistic Regression, Decision Tree, and KNN, based on a two-year case study of credit card transactions, and their accuracies were calculated [14]. Yaumil's study examined fraud perpetrators of doctors and providers through unadministered duplicating claims of insurance and medical action by data tracking, and by providing new technology [16Aziani (2021) explores how the COVID-19 pandemic has led to organized crime amongst wholesale medicine circulation between pharmaceuticals, hospitals and patients [17].Teresa's (2022) study investigates the use of value-based adjusted payments to vulnerable patients to assess performance measures of service providers and social risk factors, and reduce readmissions penalties for hospitals [18]. Guoming's (2021) study looks into the rising of medical treatment and expenses based on diagnosis-related groups of patient insurance settlement [19]. Abhishek's study on Total Interpretive Structural Modeling Analysis of Health Care Management Negligence and Waste Management using Structural Equation Modeling [20]. Steve's study on the Healthcare Sector and the Recrudescence of Cyber-Attacks, and their Impact on Privacy in Hospitals, analyzed using [21]. Mahdi's study on Supervised Learning to Detect Fraud and Misinformation in Supply Chains, using Macro-level Python Code [22]. Ammar's study on Blockchain Security, Centralization of Cryptographic Systems, and Electronic Records in the Healthcare Field [23]. Charron's study [24] examined Medicare, financial strain, and poor quality of food and shelter in older adults according to race and economic status. Zheng's study [25] used an ensemble learning algorithm and management discussion analysis to identify financial fraud, and analyzed sentiment polarity. Jenita Mary's study [26] evaluated the false positive rate for insurance fraud detection by applying k-mean clustering, Support Vector Machine (SVM), and Naive Bayes (NB) on a healthcare provider fraud detection dataset. The study clearly shows the organized crime and malpractice between hospitals, diagnostic centers, doctors, insurance, and pharmaceutical companies, involving money laundering, fraud, and extortion. This fraudulent activity has a major effect on patients' full Mediclaim and insurance companies' financial aid.

**General analysis and related crimes and prevention methods**

Detection and correlation of medical claim service provider reimbursement analysis as per study of Gaskins, 2017 as follows 1. Medical claim service provider reimbursement analysis is a process used by insurance companies and other entities to detect and identify fraudulent billing practices and other unauthorized activity. 2. The process uses data analytics to look for patterns and correlations in claims being processed for payment. 3. Medical claim service provider reimbursement analysis is one of the most effective tools available to combat healthcare fraud. 4. By using data analytics to identify potential anomalies in medical claims, healthcare organizations can reduce their risk of losing money due to fraudulent activities. 5. Medical claim service provider reimbursement analysis can also help identify billing errors and other processes that could result in overpayment or underpayment.6. By carefully monitoring claims and identifying patterns, medical claim service provider reimbursement analysis can help healthcare organizations ensure they receive the correct payments for services rendered. 7. Healthcare organizations that use medical claim service provider reimbursement analysis can also improve their internal processes and reduce the chance of future fraudulent activities. 8. Regular reviews of medical claims can help ensure that all claims are accurate and that payments are received in a timely manner. 9. Healthcare organizations can use medical claim service provider reimbursement analysis to identify common trends in fraudulent billing practices and other healthcare fraud schemes. 10. Healthcare organizations can also use medical claim service provider reimbursement analysis to develop better billing processes and procedures to prevent fraud. 11. Additionally, medical claim service provider reimbursement analysis can be used to identify areas of overpayment or underpayment that can be addressed. 12. Medical claim service provider reimbursement analysis is an important tool for healthcare organizations to detect, prevent, and reduce healthcare fraud. 13. Healthcare organizations should use medical claim service provider reimbursement analysis to identify suspicious activities and take appropriate actions to prevent future fraud. 14. Healthcare organizations can also use medical claim service provider reimbursement analysis to identify areas of potential risk and take steps to mitigate those risks. 15. Healthcare organizations should also use medical claim service provider reimbursement analysis to identify patterns of fraudulent

Medical claim service providers must comply with various laws and regulations to ensure the safety and accuracy of the claims they process. These laws and regulations vary from state to state, but typically include requirements for the provider to be licensed, bonded, and insured, as well as to maintain certain record-keeping and reporting procedures. In some cases, the provider must also comply with HIPAA regulations, which protect the privacy of patient health information. Additionally, the provider must comply with state and federal laws governing billing and reimbursement practices. Finally, the provider must comply with state and federal laws governing the use of electronic health records.

**Methodology**

The public dataset was used for this study and the methodology to predict organized fraud in health care service providers was based on regression models. The first step was to resize and normalize the dataset to train and test beneficiaries' details based on their illness and chronic conditions of various diseases and disorders. Zixuan Qin's (2022) study showed that the CNN regression model was used for the real-time and virtual control systems of data [27], and the model was used to classify both the normal (0) and chronic (1) conditions. Based on the chronic condition, the birth and death dates of the patient were calculated. The Inpatient and Outpatient case history of admission and discharge details were then merged. This merged dataset was further combined with beneficiary and health care service provider data to identify any fraudulent claims. The proportion of potential fraudulent claims was calculated using the merged data, providing insight into the number of transaction claims and money involved between the beneficiary, physicians, and diagnostic centers. An evaluation of the top reasons for suspicious money transactions between health care and diagnostic centers based on fraud and non-fraud counts was conducted. This analysis allowed for accurate predictions of the top physician malpractice with insurance companies and diagnostic centers. The relation between insurance provider claims transaction of patients and fraud cases is determined by their age. Fraud cases are more prominent in the age group above 70 years and less in the age group of 30 to 70 years. Age is thus an important factor in distinguishing between fraud and nonfraud behavior. To accurately select the features, a CNN-based feature selection was used. Clues of fraud and abuse were identified by employing sparse matrix and grouping-based similarity calculations. The sparse matrix is used for compression techniques to store elements in memory more efficiently by replacing zero values with fewer memory footprints. Similarly, based grouping is used to split a dataset into distinct groups based on their similar constraints and variance. This technique is often used to group records of clinical procedures, diagnostic, and insurance provider codes. Principal component analysis (PCA) is used to reduce the data set into the lower dimension and Auto Encoders (AE) is used to reduce the original dimensionality of the data set and generate the compressed representation of the original data. Both PCA and AE are used for the feature extraction which are further used for the classification Random Forests consist of multiple random decision trees that are used to classify datasets based on their features and convert weak learners into strong learners during training. To extract the variance of features, Principal Component Analysis (PCA) and Autoencoders (AE) are utilized. The proposed Auto-Encoder model is primarily used to extract more features by thresholding the reconstruction error on a merged dataset. The errors on the individual attributes of fraud and non-fraud data are combined to create the target class, and then all the attribute errors are concatenated with the target class based on their variance. The overall prediction is based on performance metrics such as curves, correlation of residual error, threshold, error loss, and the ROC curve.

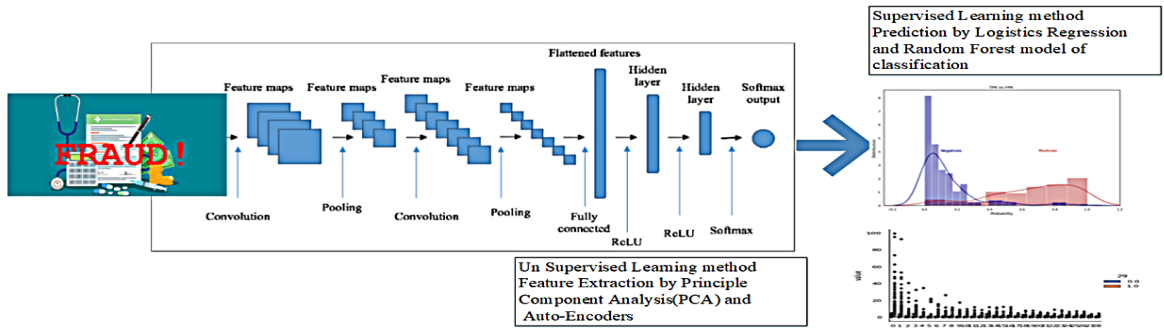

Figure 1. The architecture with supervised and unsupervised classifiers

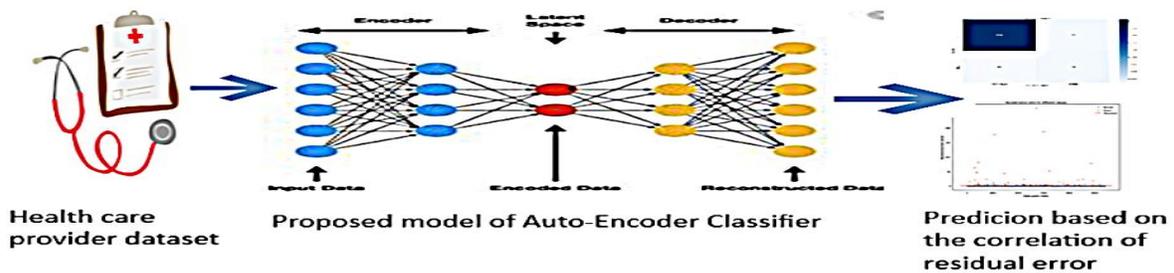

Figure 2. The proposed model

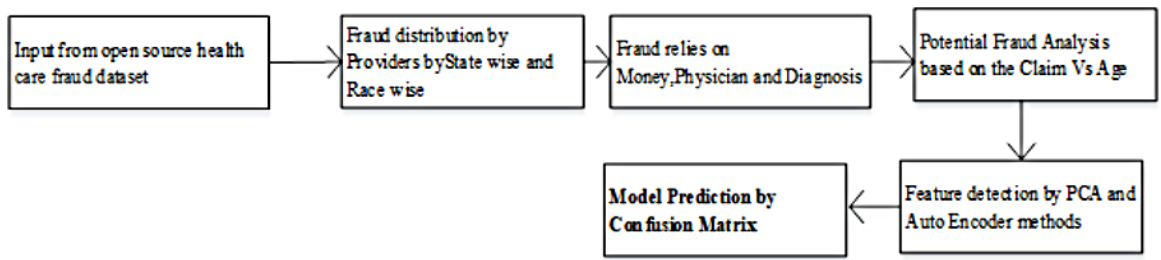

Figure 3. The proposed methodology of model prediction by confusion matrix

**Logistics Regression model**

The prediction is based on linear dependent and independent variables. According to Atari (2019), logistic regression is an effective model for determining health insurance ownership (yes/no) based on age and gender [28]. Similarly, Schober (2021) states that logistic regression can be used to estimate the relationship between one or more independent variables and a binary (dichotomous) outcome variable [29]. The model is trained and validated, then the ROC analysis is conducted by measuring truth positive (precision) against false positive (recall). The probability threshold can be adjusted to change the threshold between 50-50 at the trade-off level, which optimizes the f1-measure metrics. However, the accuracy of the model is decreased when the cut-off is modified.

**Random Forest model**

The Random Forest model is used to predict health care variables and is shown to be more accurate than other models [30]. It utilizes non-linearity between variables to group a subset of decision trees and identify predictors through the outcome of a discriminate result of poor versus good (AUC = 0.707) [31]. The model performs both training and validation and then uses ROC analysis to evaluate truth positive (precision) versus false positive (recall). The features are extracted based on claim, reimbursement, diagnosis, and physicians' charges. The model is also used to predict unseen data.

**Principle component Analysis (PCA)**

This is the unsupervised learning method used for feature extraction. It helps to standardize scalar data with the maximum number of variables, allowing for the identification of linear and non-linear variance in order to extract features that may indicate potential fraud information. Multicollinearity in the models is addressed by using Principal Components Analysis (PCA) to produce linearly uncorrelated components [32]. PCA has been used for clinical intervention, such as for evidence-based decision-aids [33], as well as for data pre-processing, such as for feature extraction and dimension reduction.

**Proposed model of Auto-Encoders (AE)**

The model is used for learning generative information through encoding and dimensionality reduction methods. It splits the features into a reduced representation and reconstruction error, and then generates outputs as close as possible to the original input. The model is used to learn the patterns from non-fraudulent data and then uses the reconstruction error threshold on fraudulent information to predict the target class. The model is a sequential network layer with a dropout (0.2) layer to add noise to the data and a dense layer to encode patterns via dimensionality reduction. The model has a batch size of 32 and a learning rate of 100 epochs. The model predicts the output based on the mean square error as a metric, with each neuron acting as a linear regression method. It computes and combines errors from fraud and non-fraud information based on the error plot. A fixed threshold is used for the classification of fraud and non-fraud information. The shape of the target is expanded for merging and concatenating all the attribute errors with the target class. The percentile distribution of absolute errors is then calculated. Different threshold values are used for the reconstruction of error, and the evaluation result shows improved performance of the model. The Auto encoders have two hidden layers added in order to calculate the reconstruction error for both normal and fraudulent data. The graphical representation demonstrates the ROC analysis of true class, precision vs. recall metrics, threshold vs. recall, and recall vs. threshold. The model is able to determine the prediction class based on the error threshold. Dual Autoencoders are employed to address the issue of imbalanced classes when embedded in the training data [34]. Autoencoders are also highly effective when used for predictive analysis, such as diagnosing faults and determining the useful life of components [35].

**Performance Metrics Evaluation**

Performance metrics was calculated based on the true positive, negative and false positive, negative rate based on the confusion matrix.

The area under the curve (AUC) ranges between 0 and 1, with an AUC of 0.0 indicating wrong prediction and an AUC of 1.0 indicating perfect prediction. The Receiver Operating Characteristic (ROC) is used to predict a dichotomous outcome by comparing sensitivity against specificity. By measuring the area under the ROC curve, it is possible to evaluate and compare the performance of different classifiers. The Cohen Kappa statistic is used to compare the observed accuracy to the expected accuracy, and to evaluate the performance of different classifiers.

**Results and Discussion**

The percentage distribution of the fraud class "No" is 90.64695 and the distribution of the fraud class "Yes" is 9.35305. Over 80% of beneficiaries are of the same race. Figure 4 illustrates that the top diagnoses (in terms of money involved) are 4019, 4011, and 2724 and the distribution of fraud and non-fraud counts reveals suspicious transactions associated with them. Figures 4 illustrate the performance of the logistic regression model through precision and recall curves, respectively. Figure 5 depicts the performance of the random forest model through true positive and false positive rates. Figure 6 Auto Encoder prediction based on reconstruction error threshold of fraud data with ROC curve and confusion matrix

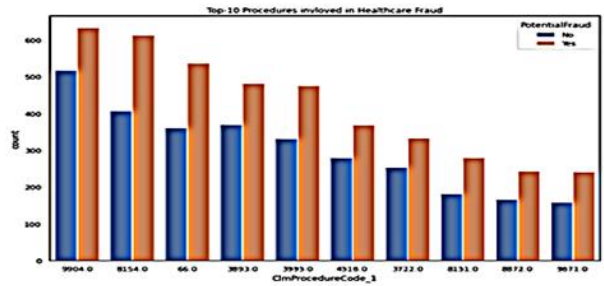

**Figure 4.** Distribution of fraud and non-fraud count shows suspicious transactions involved in them.

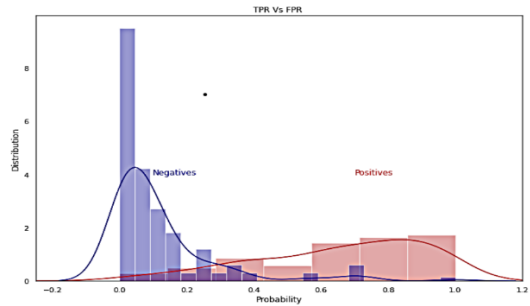

**Figure 5.** Logistic Regression trade off metrics of Precision (TPR) and Recall (FPR)

Logistic regression is a supervised learning algorithm that is used for classification problems. It is a type of regression analysis that predicts the probability of a certain event occurring. The trade-off between precision (TPR) and recall (FPR) is a measure of how accurately the model is predicting the outcome. Precision is a measure of how many of the predicted positive outcomes are actually correct. Recall is a measure of how many of the actual positive outcomes are being correctly identified. As precision increases, recall decreases and vice versa. The ideal trade-off between precision and recall is a point where both metrics are maximized. This can be seen in Figure 6 which shows a graph of precision (TPR) and recall (FPR) against a logistic regression model. To visualizes the performance of a logistic regression model on a dataset. The ROC curve shows the model's true positive rate (TPR) against its false positive rate (FPR). The higher the TPR is, the better the model is at correctly classifying positive cases. The precision vs recall graph shows the precision (how often the model correctly predicts positive cases) versus the recall (how often the model correctly identifies all positive cases). The false positive cases graph shows the number of false positives that the model generates.

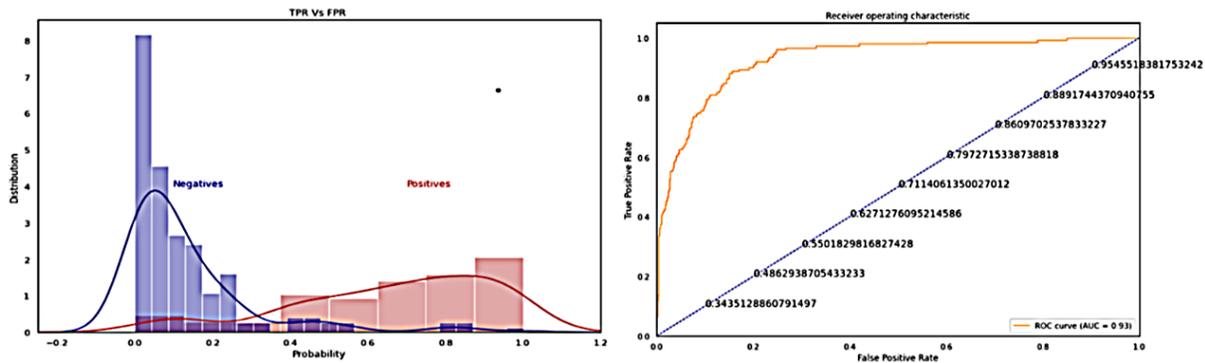

**Figure 6.** Random forest model performance metrics of TPR vs FPR and ROC curve

The figure is a visualization of the performance metrics of a Random Forest model. The graph plots the true positive rate (TPR) on the y-axis against the false positive rate (FPR) on the x-axis. The graph also shows the ROC (receiver operating characteristic) curve which is a measure of the model's performance at different thresholds. The area under the curve (AUC) is a measure of accuracy and the closer the AUC is to 1, the better the model's performance is. The graph can be used to determine the best threshold for the model, which is the point at which the TPR and FPR are maximized. The autoencoder is a type of artificial neural network that is used to identify anomalies in data. In this process, non-fraud data is first used to train the autoencoder. The autoencoder then reconstructs the non-fraud data and the reconstruction error is used as a threshold for identifying fraud data. Finally, the reconstruction error of the fraud data is compared to the threshold to determine whether or not it is fraudulent.

An Auto Encoder (AE) is an unsupervised neural network used for learning efficient data representations. It is composed of an encoder, which maps the input data to a latent space, and a decoder, which reconstructs the input data from the latent space. The AE is trained to minimize the difference between the input and reconstructed data. In this figure, the AE is used to detect fraudulent data by training it on a dataset of known fraudulent data. The AE is then used to predict whether new data is fraudulent or not based on the error threshold for reconstruction. The results of the prediction are shown on a ROC curve and a confusion matrix. The ROC curve plots the true positive rate against the false positive rate, and the confusion matrix displays the model's performance in terms of true positives, false positives, true negatives, and false negatives. The higher the area under the ROC curve, the better the model is at distinguishing between fraudulent and non-fraudulent data. The confusion matrix can be used to determine the precision, recall, and F1 score of the model.

**Dataset**
The open-source dataset in Kaggle added with some private data

**Conclusion**

The anomaly detection algorithms are used to detect the outliers and the out-of-distribution data points. The supervised learning algorithms are used to classify the data into different classes and the unsupervised learning algorithms are used to cluster the data into different groups. The fraud detection techniques are used to predict the fraudulent activities and the machine learning algorithms are used to analyze the data and make the better decisions based on the patterns. The fraud detection techniques are used to detect the anomalies and the non-compliant activities. The supervised learning algorithms are used to classify the data into different classes and the unsupervised learning algorithms are used to cluster the data into different groups. The fraud detection techniques are used to identify the fraudulent activities and the patterns of the data. The data mining techniques are used to analyze the data and detect the patterns and anomalies. The data mining techniques are used to identify the underlying patterns and the outliers. The data mining techniques are used to improve the accuracy of the predictions and the decisions. The analytics techniques are used to analyze the data and detect the patterns. The analytics techniques are used to identify the underlying trends and the outliers. The analytics techniques are used to improve the accuracy of the predictions and the decisions. The fraud detection techniques are used to identify the fraudulent activities and the patterns. The fraud detection techniques are used to detect the anomalies and the non-compliant activities.